\documentclass[runningheads]{llncs}
\usepackage{graphicx}
\usepackage{comment}
\usepackage{amsmath,amssymb} % define this before the line numbering.
\usepackage{color}
\usepackage{algorithm}
\usepackage{algorithmic}
\usepackage{wrapfig}

\usepackage{multirow}
\usepackage{multicol}
\usepackage{mathtools}
\usepackage{url}

\usepackage{xr}
\makeatletter
\newcommand*{\addFileDependency}[1]{
  \typeout{(#1)}
  \@addtofilelist{#1}
  \IfFileExists{#1}{}{\typeout{No file #1.}}
}
\makeatother

\newcommand*{\myexternaldocument}[1]{
    \externaldocument{#1}
    \addFileDependency{#1.tex}
    \addFileDependency{#1.aux}
}
\myexternaldocument{4889-supp}

\begin{document}
% \linenumbers
\pagestyle{headings}
\mainmatter
\def\ECCVSubNumber{4889}  % Insert your submission number here

\title{Improving Query Efficiency of Black-box Adversarial Attack} % Replace with your title

\begin{comment}
\titlerunning{ECCV-20 submission ID \ECCVSubNumber} 
\authorrunning{ECCV-20 submission ID \ECCVSubNumber} 
\author{Anonymous ECCV submission}
\institute{Paper ID \ECCVSubNumber}
\end{comment}
%******************

% \begin{comment}
\titlerunning{Improving Query Efficiency of Black-box Adversarial Attack}
% \textsuperscript{$\diamond$}
\author{Yang Bai \inst{1,4,}\textsuperscript{$\star$} \and
Yuyuan Zeng\inst{2,4,}\textsuperscript{$\star$} \and \\ Yong Jiang\inst{1,2,4,}\textsuperscript{$\dagger$} \and Yisen Wang\inst{3,}\textsuperscript{$\dagger$} \and Shu-Tao Xia\inst{2, 4} \and Weiwei Guo\inst{5}
}
\authorrunning{Y. Bai et al.}
\institute{Tsinghua Berkeley Shenzhen Institute, Tsinghua University \and
Tsinghua Shenzhen International Graduate School, Tsinghua University \and 
Shanghai Jiao Tong University \and
PCL Research Center of Networks and Communications, Peng Cheng Laboratory \and 
vivo AI Lab}
% \end{comment}
%******************
\maketitle

\renewcommand{\thefootnote}{\fnsymbol{footnote}} 
\footnotetext[1]{Equal contribution. (\{y-bai17, zengyy19\}@mails.tsinghua.edu.cn)}
\footnotetext[4]{Corresponding authors: Yisen Wang (eewangyisen@gmail.com) and Yong Jiang (jiangy@sz.tsinghua.edu.cn). } 

\begin{abstract}
Deep neural networks (DNNs) have demonstrated excellent performance on various tasks, however they are under the risk of adversarial examples that can be easily generated when the target model is accessible to an attacker (white-box setting). As plenty of machine learning models have been deployed via online services that only provide query outputs from inaccessible models (\textit{e.g.}, Google Cloud Vision API2), black-box adversarial attacks (inaccessible target model) are of critical security concerns in practice rather than white-box ones. However, existing query-based black-box adversarial attacks often require excessive model queries to maintain a high attack success rate. Therefore, in order to improve query efficiency, we explore the distribution of adversarial examples around benign inputs with the help of image structure information characterized by a Neural Process, and propose a Neural Process based black-box adversarial attack (NP-Attack) in this paper. Extensive experiments show that NP-Attack could greatly decrease the query counts under the black-box setting. Code is available at \url{https://github.com/Sandy-Zeng/NPAttack}.

\keywords{Black-box Adversarial Attack, Adversarial Distribution, Query Efficiency, Neural Process}
\end{abstract}

\section{Introduction}
Deep neural networks (DNNs) have been deployed on many real-world complex tasks and demonstrated excellent performance, such as computer vision \cite{he2016deep}, speech recognition \cite{wang2017residual}, and natural language processing \cite{devlin2019bert}. However, DNNs are found vulnerable to adversarial examples, \textit{i.e.}, DNNs will make incorrect predictions confidently when inputs are added with some well designed imperceptible perturbations \cite{goodfellow2014explaining}, thus various adversarial defense methods are proposed \cite{bai2019hilbert,wang2019convergence,wang2019improving}. Adversarial examples can be crafted following either a white-box setting (the adversary has full access to the target model) or a black-box setting (the adversary has no information of the target model). White-box methods such as Fast Gradient Sign Method (FGSM) \cite{goodfellow2014explaining}, Projected Gradient Decent (PGD) \cite{madry2017towards}, Carlini \& Wagner (CW) \cite{carlini2017towards} and other universal attacks \cite{Moosavi-Dezfooli_2017_CVPR,liu2020self} only pose limited threats to DNN models which are usually kept secret in practice. Meanwhile transferability-based black-box attacks, \textit{e.g.}, momentum boosting \cite{dong2018boosting} and skip gradient \cite{wu2020skip}, need to train a surrogate model separately and only obtain a moderate attack success rate. 

As modern machine learning systems, \textit{e.g.}, Google Cloud Vision API2 (\url{https://cloud.google.com/vision/}) and Google Photos3 (\url{https://photos.google.com/}), are often provided as a kind of service, one common scenario is that we can query the system in a number of times and get the output results \cite{bai2020targeted}. Based on this, \textit{query}-based black-box attacks that directly generate adversarial examples on the target model are proposed such as ZOO \cite{chen2017zoo}. These query-based methods could bring almost 100\% attack success rates while their query complexity is quite high (not acceptable). Therefore, how to significantly reduce the query complexity while maintaining the attack success rate simultaneously is still an open problem. There are several existing work to reduce the query complexity. For examples, AutoZOOM \cite{tu2019autozoom} compresses the attack dimension while QL \cite{ilyas2018black} adopts an efficient gradient estimation strategy. Different from the above example-wise adversarial example generation, $\mathcal{N}$Attack \cite{li2019nattack} proposes to model the adversarial distribution and sample from it to generate adversarial examples, which indeed reduces the query counts. However, the distribution is based on the simple pixel-wise mapping functions (\textit{e.g.}, $tanh$), which is the bottleneck of its query complexity. 

\begin{figure}[!t]
    \centering
    \includegraphics[width=12cm]{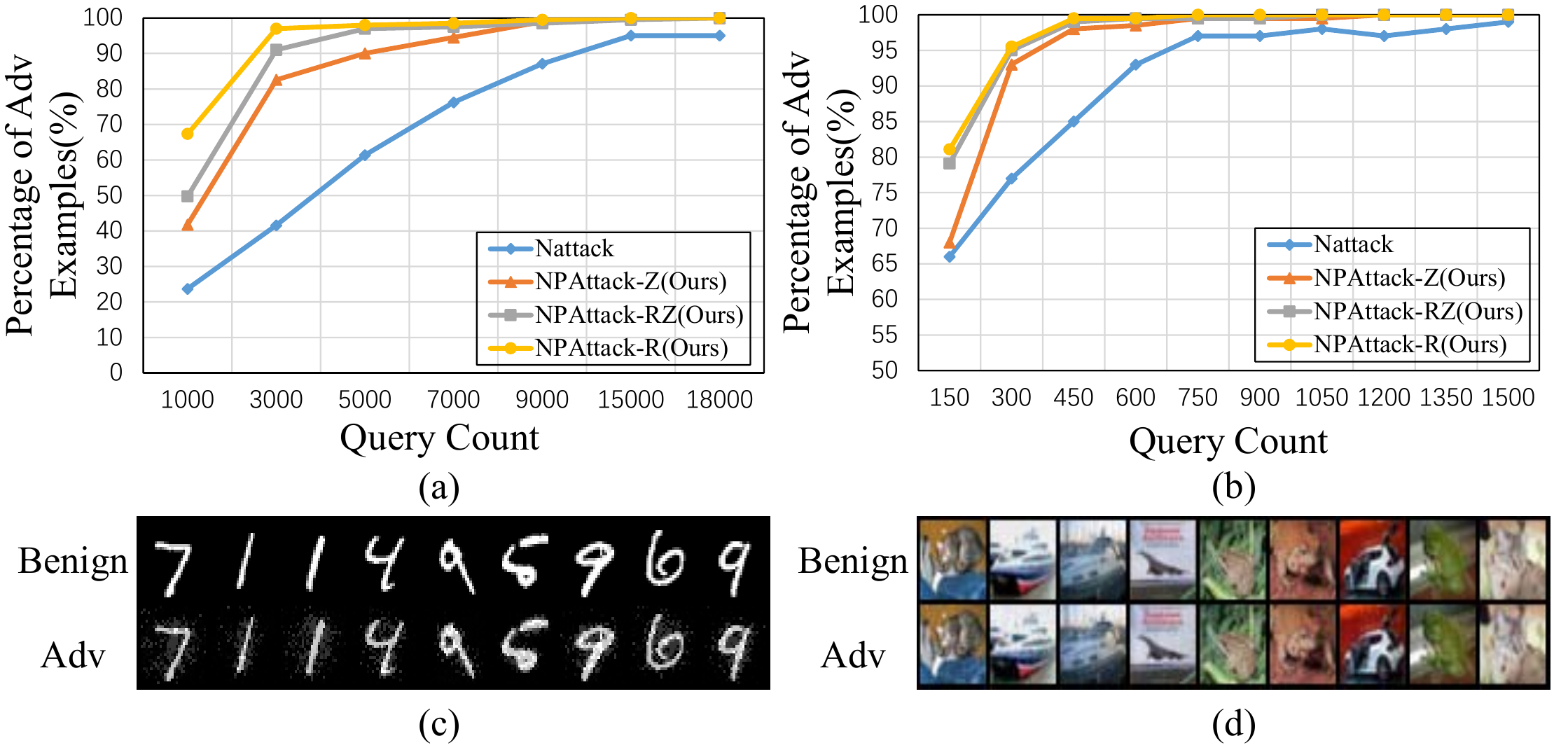}
    \caption{Comparison of two distribution-based black-box attacks: $\mathcal{N}$Attack and our proposed NP-Attack. Percentage of adversarial examples with different query counts generated by $\mathcal{N}$Attack and NP-Attack on MNIST (a) and CIFAR10 (b), and the corresponding NP-Attack generated adversarial examples and benign examples of MNIST (c) and CIFAR10 (d). It is demonstrated that NP-Attack is more query efficient compared to $\mathcal{N}$Attack when generating adversarial examples.}
    \label{fig:intro}
\end{figure}

Inspired by the above observations, in this paper, we introduce the structure information of the image into consideration to further reduce the required query counts when modeling the distribution of adversarial examples. To be specific, the structure is characterized by a Neural Process (NP) \cite{Garnelo2018Neural}, an efficient auto-encoder method to model a distribution over regression functions with a deterministic variable focusing on local information and a latent variable focusing on global information. 
Based on NP, we then propose a Neural Process based black-box attack, named NP-Attack. NP can be pre-trained on benign examples, and then used to reconstruct adversarial examples through optimizing above mentioned variables, which is definitely efficient than previous pixel-wise operations in $\mathcal{N}$Attack. As a proof-of-concept, we conduct an experiment to compare these two distribution-based attacks: $\mathcal{N}$Attack and NP-Attack. Both attacks are conducted by limiting the same maximal query count, and the maximal distortion of $L_\infty$ is set to 0.05 for CIFAR10 and 0.2 for MNIST. The percentage of adversarial examples returned at the maximal query count is plotted in Figure \ref{fig:intro}. We can see that the adversarial distribution optimized in NP-Attack (three variants) contains a higher percentage of adversarial examples under the same query counts compared to $\mathcal{N}$Attack, which illustrates NP-Attack is more efficient in optimizing and modeling the adversarial distributions. Figure \ref{fig:intro} also gives some examples of adversarial examples generated in our NP-Attack, which are also visual integrity. Our main contributions could be summarized as follows:

\begin{itemize}
\item We propose a distribution based black-box attack, Neural Process based black-box attack (NP-Attack), which uses the image structure information for modeling adversarial distributions and reduces the required query counts. 

\item NP-Attack has several optimization variants due to the variables in NP. The optimization on deterministic variable focuses more on the local information, while optimization on latent variable focuses more on the global information. The different optimization variants have different effects on the location of adversarial perturbations, which brings more flexibility for NP-Attack.

\item Extensive experiments demonstrate the superiority of our proposed NP-Attack. On both untargeted and targeted attacks, NP-Attack greatly reduces the needed query counts under the same attack success rate and distortion, compared with the state-of-the-art query-based black-box attacks.

\end{itemize}

\section{Related Work}

Existing black-box attacks can be categorized into two groups: 1) transferability-based method that transfers from attacking a surrogate model; and 2) query-based method that directly generates adversarial examples on the target model. 

For transferability-based black-box attacks, adversarial examples are crafted on a surrogate model then applied to attack the target model. There are several techniques to improve the transferability of black-box attacks. For example, Momentum Iterative boosting (MI) \cite{dong2018boosting} incorporates a momentum term into the gradient to boost the transferability. Diverse Input (DI) \cite{xie2019improving} proposes to craft adversarial examples using gradient with respect to the randomly-transformed input example. Skip Gradient Method (SGM) \cite{wu2020skip} uses more gradients from the skip connections rather than the residual modules via a decay factor to craft adversarial examples with high transferability. However, they usually cannot obtain the 100\% attack success rate, which is closely restricted by the dependency between the surrogate model and the target model. 

For query-based black-box attacks, they could be further classified as decision-based (query results are one-hot labels) \cite{chen2020boosting} or score-based ones (query results are scores). The score-based attacks are investigated in this paper. This kind of method estimates the gradient of the target model via a large number of queries, which is then used to generate adversarial examples. ZOO \cite{chen2017zoo} explores gradient estimation methods by querying the target model as an oracle. They use zeroth-order stochastic coordinate descent along with dimension reduction, hierarchical attack and importance sampling techniques, to directly estimate the gradients of the targeted model for generating adversarial examples. However, ZOO requires numerous queries to estimate the gradients with respect to all pixels. Further, AutoZOOM \cite{tu2019autozoom} operates the gradient estimation in latent space, using an offline pre-trained auto-encoder or a bilinear mapping function to compress the attack dimension. It then applies an adaptive random gradient estimation strategy to balance query counts and distortion, which improves the query efficiency by a great deal. Meta Attack \cite{du2019query} pre-trains a meta attacker model to estimate the black-box gradient, which efficiently reduces the query counts. Beyond zeroth-order optimization-based approaches, QL \cite{ilyas2018black} proposes to use a Natural Evolution Strategy (NES) \cite{wierstra2008natural} to enhance query efficiency. Bandits \cite{ilyas2018prior} further introduces time and data priors under NES. $\mathcal{N}$Attack \cite{li2019nattack} is another kind of black-box attack that explicitly models the adversarial distribution with a Gaussian distribution. The adversarial attack is hence formalized as an optimization problem, which searches the Gaussian distribution under the guidance of increasing the attack success rate of target models. 

Following the general NES structure, we explore the adversarial distribution in a more efficient way using some high-level information of images in this paper.

\section{Proposed Neural Process-based Black-box Attack}
In this section, we propose a Neural Process-based Black-box Attack (NP-Attack) with significantly reduced query counts. NP-Attack models the distribution of adversarial examples efficiently by a Neural Process that utilizes the high-level structure information of images rather than the pixel-level information.

\subsection{Preliminaries of Neural Process}
Neural Process (NP) \cite{Garnelo2018Neural} is a combination of the best from neural networks and Gaussian Process \cite{Matthews2018Gaussian,Wistuba2018Scalable}, which could efficiently estimate the uncertainty in the predictions. It could be expanded into an attentive version, called Attentive Neural Process (ANP) \cite{kim2019attentive,vaswani2017attention}, which is applied in our NP-Attack\footnote{We still use NP in the following without ambiguity}. 

\begin{figure}[tb]
    \centering
    \includegraphics[width=10cm,height=3cm]{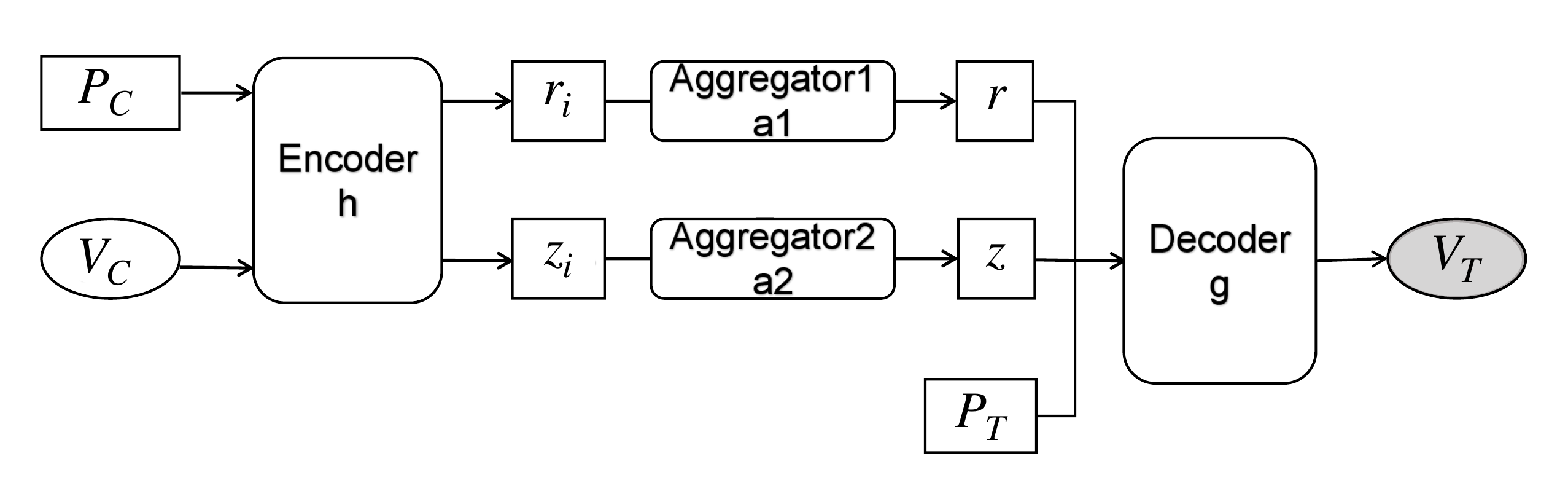}
    \caption{The structure of NP model trained on pixels of one image. Here, $V_{C}$ and $V_{T}$ are the context and target pixel values, then $P_{C}$ and $P_{T}$ are the corresponding pixel positions. $z_i$ and $r_i$ are pixel-wise latent variables and deterministic variables.}
    \label{fig_NP}
    % \vspace{-0.5cm}
\end{figure}

As pixel values in one image subject to a Gaussian Process, NP is applied to reconstruct images by predicting their pixel values. 
%In this paper, we train the NP model with pixel values of one image. 
As shown in Figure \ref{fig_NP}, the structure of NP consists of the following three parts: 
1) \textit{Encoder $h$}, whose inputs are context pixel pairs concatenating pixel positions and values; 
2) \textit{Aggregator $a$}, which summarizes the outputs of the encoder as latent variable $z$ from a Gaussian distribution $\mathcal{N}(\mu, \sigma^2)$ and deterministic variable $r$;
3) \textit{Decoder $g$}, which takes the sampled latent variable $z$, deterministic variable $r$ and target pixel position $P_{T}$ as inputs and predicts the target pixel value $V_{T}$. 
Overall, NP models the distributions as:
\begin{equation}
\centering
    p \left( V_{{T}}{ \left| {P_{{C}},P_{{T}},V_{{C}} \left)={ \int {p \left( V_{{T}}{ \left| {P_{{T}},r,z \left) q \left( z{ \left| {s_{{C}} \left) dz, \right. }\right. }\right. \right. }\right. }\right. }}\right. }\right. }\right.
\label{np_distribution_equation}
\end{equation}
Here $P$, $V$ represents pixel position and value, and $C$, $T$ are random subsets of context and target, $s_C$ is the distribution modeled over ($P_C$, $V_C$). 
In Eq. \ref{np_distribution_equation}, Latent variable $z$ accounts for uncertainty in the predictions of $V_T$ for observed ($P_C$, $V_C$), while $r$ is calculated by a deterministic function which aggregates ($P_C$, $V_C$) into a finite dimensional representation with permutation invariance in $C$. The interpretation of the latent path is that $z$ gives rise to correlations in the marginal distribution of the target predictions $V_T$ , modeling the global structure of the stochastic process realization, whereas the deterministic path $r$ models the fine-grained local structure. 

\subsection{Pre-training of Neural Process}
\label{sec:distribution}

The NP model in NP-Attack is pre-trained on the pixels of given benign images by maximizing the following Evidence Lower Bound (ELBO)\footnote{The implementation details of ANP are shown in the Appendix \ref{appendix:structure}}:
\begin{equation}
\label{np_eblo_equation}
\begin{split}
\log p \left( V_{{T}}{ \left| {P_{{T}},P_{{C}},V_{{C}}}\right. } \right) & \ge 
{\rm E} \mathop{{}}\nolimits_{{q \left( z{ \left| {\left. s_{{T}} \right) }\right. }\right. }}{ \left[ {\log p \left( V_{{T}}{ \left| {\left. P_{{T}},r,z \right) }\right. }\right. } \right] }  -D\mathop{{}}\nolimits_{{KL}} \left( q \left( z{ \left| {s_{{T}} \left) { \left\Vert {q \left( z{ \left| {s_{{C}} \left)  \right). }\right. }\right. }\right. }\right. }\right. }\right. \right.  
\end{split}
\end{equation}
Like $s_C$ defined above, $s_T$ represents the distribution modeled over ($P_T,V_T$). To be specific, the $s_C$ and $s_T$ are both the whole pixels in one image during the pre-training process in NP-Attack. Once trained, NP shares the same encoder and decoder on the same image set even on different selected images, varying on latent variables and deterministic variables corresponding to different images. Note that such pre-training process is independent from the main attack part on the benign examples, which means the encoder and decoder are fixed after pre-training and could keep some structure information of images. 
%In fact, NP defined above could be pre-trained on benign images or adversarial ones. The adversarial pre-trained one adopts adversarial examples generated on some substitute model in advance. As adversarial examples show transferability, adversarial pre-trained NP could have a bias on adversarial examples thus be more efficient in black-box attack later. However as we would like to compare our method with those absolutely query based attack techniques, we just adopt the benign pre-trained NP, guaranteeing a fair comparison without any substitute model. 

The pre-trained NP models a distribution by variables $z$ and $r$, where $z$ is sampled from a Gaussian distribution $\mathcal{N}(\mu, \sigma^2)$, thus the latent variables and deterministic variable are in fact $(\mu, \sigma)$ and $r$. As NP models the pixel-wise distribution in one image, its structure information is kept in such NP. Adversarial examples and benign examples are imperceptible to human eyes, sharing a similar visual structure. Thus we utilize the pre-trained NP and find that adversarial examples could be reconstructed successfully by optimizing the above mentioned variables for a new distribution and sampling from this optimized distribution. 

\subsection{Overview of the Proposed NP-Attack} 
Equipped with the pre-trained NP, we can propose the distribution-based NP-Attack. The key idea of distribution-based attack is to model the adversarial distribution around the small region of one natural example $x$, such that a sample drawn from this distribution is likely an adversarial example. Compared to previous distribution-based $\mathcal{N}$Attack, the difference lies on the method of modeling the adversarial distribution. $\mathcal{N}$Attack models such adversarial distribution just on pixel level with a Gaussian distribution focusing on pixels independently, not considering any other information on the structure of pixel values in one image. However, our proposed NP-Attack utilizes the decoder of a pre-trained NP model and only optimizes latent variables or deterministic variables, in which case the fixed decoder could hold some structure information of pixels in one image, improving the optimization efficiency in modeling adversarial distributions.

Specifically, the objective function $L$ in our NP-Attack is defined on $S$, an intersection of a latent region modeled by NP and a $L_p$-ball centered around the benign example $x$, \textit{i.e.}, $S = S_{p}(x) \bigcap \text{NP}(x)$. 
Given $l$ as the loss defined on a given example point, the objective $L$ with regard to adversarial distribution in our optimization criterion is:
\begin{equation}
L((\mu, \sigma), r|x) := \int_{x-\epsilon}^{x+\epsilon} l(x_{rec})g(x_{rec}|(\mu, \sigma), r)h((\mu, \sigma), r|x)dx_{rec}
\end{equation}
where $g$ and $h$ are the decoder and encoder of pre-trained NP, $\mu, \sigma, r$ denote for the variables to be optimized in NP-Attack, $x_{rec}$ denotes the image reconstructed by NP, $\epsilon$ denotes the $L_p$ ball restriction. By optimizing on such objective $L$, we could achieve our aim to model the latent manifold of adversarial examples.

As the target model could return scores in each query, the original loss function is defined as:
\begin{equation}
\label{eq:loss}
l(x):=\left\{
\begin{aligned}
\max(0,\max_{c \neq y} \log F(x)_{c}-\log F(x)_{y}) & , & \text{targeted}, \\
\max(0,\log F(x)_{y}-\max_{c \neq y} \log F(x)_{c}) & , & \text{untargeted}.
\end{aligned}
\right.
\end{equation}
where $F(x)$ denotes the softmax outputs, $y$ is the true label in untargeted attack or the target label in targeted attack, and $c$ is other labels except $y$.

In summary, the procedures on modeling and sampling from distribution in NP-Attack are shown in Algorithm \ref{alg:NP-Attack}: 1) Feed a benign example $x$, and compute $r$ and $(\mu, \sigma)$ from the \textit{Encoder $h$} of NP; 2) Sample $z$ from $\mathcal{N}(\mu, \sigma^2)$; and 3) Reconstruct $x_{rec}$ with the optimized ($r$, $z$) from the \textit{Decoder $g$} of NP (shown in the following Section \ref{sec: optimization}), then project $x_{rec}$ back into the $L_p$-ball centered at $x$.

\begin{algorithm}[!h]
\renewcommand{\algorithmicrequire}{\textbf{Input:}}
\renewcommand{\algorithmicensure}{\textbf{Output:}}
\caption{NP-Attack}
\label{alg:NP-Attack}
\begin{algorithmic}[1]
   \REQUIRE natural image $x$, label $y$, target neural network $F$, pre-trained NP model (\textit{Encoder $h$}, \textit{Decoder $g$}, \textit{Aggregator $a$}), maximal optimization iteration $T$, sample size $b$, projecting function $P$, learning rate $\eta$
   \STATE Compute the variables from \textit{Encoder h} on image $x$: $\mathcal{N}(\mu, \sigma^2), r \leftarrow h(x)$
   \FOR{$t=0$ {\bfseries to} $T-1$}
   \IF {F(x) $\neq$ y (untargeted) or F(x) = y (targeted)}
   \STATE attack success.
   \ELSE
   \STATE Sample perturbations from Gaussian distribution: $p_i \thicksim \mathcal{N}(0,I), i=1,2,...,b.$
   \STATE Add the perturbation $p_i$ in NP-Attack-R/Z/RZ, specifically on $\mu$ or $r$ or both,\\
$
\left\{
\begin{aligned}
& z_i \thicksim \mathcal{N}(\mu, \sigma^2), r_i=r + p_i \sigma , &\text{NP-Attack-R}, \\
& z_i \thicksim \mathcal{N}(\mu+p_i \sigma, \sigma^2), r_i=r , &\text{NP-Attack-Z}, \\
& z_i \thicksim \mathcal{N}(\mu+p_i \sigma, \sigma^2), r_i=r + p_i \sigma , &\text{NP-Attack-RZ}.
\end{aligned}
\right.
$
   \STATE Reconstruct the image from \textit{Decoder g}, and use project function $P$ to restrict its maximal distortion: $x_i = P(g(z_i, r_i)).$
   \STATE Compute the losses of these reconstructed image series $x_i$ under targeted or untargeted setting, \\
$
\label{loss}
l_{i}=\left\{
\begin{aligned}
\max(0,\max_{c \neq y} \log F(x_i)_{c}-\log F(x_i)_{y}) & , &\text{targeted}, \\
\max(0,\log F(x_i)_{y}-\max_{c \neq y} \log F(x_i)_{c}) & , &\text{untargeted}.
\end{aligned}
\right.
$
   \\
   The corresponding loss $l_i = l_i - mean(l).$
   \STATE Update $\mu$ or $r$ or both as optimization:\\
$
\left\{
\begin{aligned}
& r_{t+1} \leftarrow r_t - \frac{\eta}{b\sigma} \sum_{i=1}^{b} l_i p_i , &\text{NP-Attack-R}, \\
& \mu_{t+1} \leftarrow \mu_t - \frac{\eta}{b\sigma} \sum_{i=1}^{b} l_i p_i , &\text{NP-Attack-Z}, \\
& \mu_{t+1} \leftarrow \mu_t - \frac{\eta}{b\sigma} \sum_{i=1}^{b} l_i p_i, r_{t+1} \leftarrow r_t - \frac{\eta}{b\sigma} \sum_{i=1}^{b} l_i p_i , &\text{NP-Attack-RZ}.
\end{aligned}
\right.
$
   \ENDIF
   \ENDFOR
\end{algorithmic}
\end{algorithm}

\subsection{Optimization of NP-Attack}
\label{sec: optimization}
As the reconstructed image $x_{rec}$ is mainly dependent on the sampled $r$, $z$ and pre-trained \textit{Decoder $g$} of NP, there are three optimization options: NP-Attack-R/Z/RZ corresponding to the optimized variables $r$, $z$ or both. As the three branches share the similar optimization function, we take \textbf{NP-Attack-R} as an example here, and other optimization options are shown in the Appendix \ref{NPattack_branches}.

In this case, $r$ is a variable in NP, cooperating with $z$ to reconstruct an image. The difference of $r$ from $z$ is that $r$ is a deterministic variable. That is, $z$ shares the same distribution in one image for all pixels, while $r$ is independent for different pixels thus with better freedom. As adversarial perturbations are computed and added on pixels, the optimization on $r$ might be more reasonable. The benefit of optimizing $r$ than $\mathcal{N}$Attack is that $r$ is in the latent space with a larger latent dimension (128 dimension in our NP-Attack) for each pixel, which owns more capacity to perturb. 

We define the loss $L$ on some search distribution $\Pi (R\mid r)$. Omitting $z$ while inheriting \textit{Decoder $g$} from the pre-trained NP, the estimated loss function in our NP-Attack could be considered as: $L = \mathbb{E}_{\Pi (R \mid r)}l(g(R)).$
To simplify the optimization, we give $R \thicksim \mathcal{N}(r, \sigma'^{2})$, where $\sigma'$ is a hyper-parameter. Then the optimization of $r$ could be implemented by adding some random Gaussian noises and using NES. The optimization function could be computed as:
\begin{equation}
    r_{t+1} \leftarrow r_t - \frac{\eta}{b} \sum_{i=1}^{b} l(g(R_{i})) \nabla_{r_t} \log \mathcal{N}(R_{i} \mid r_t, \sigma'^2),
\end{equation}
where $b$ is the batch size. Given that $R_i = r_t + p_i \sigma'$, where $p_i$ is sampled from the standard Gaussian distribution $\mathcal{N}(0,I)$, and $\sigma'$ is a hyper-parameter, not essentially equal to $\sigma$ in $z$,
$\nabla_{r_t} \log \mathcal{N}(R_{i} \mid r_t, \sigma'^2) \varpropto \sigma'^{-1} p_i.$

\begin{figure}[!t]
    \centering
    \includegraphics[width=7cm,height=3cm]{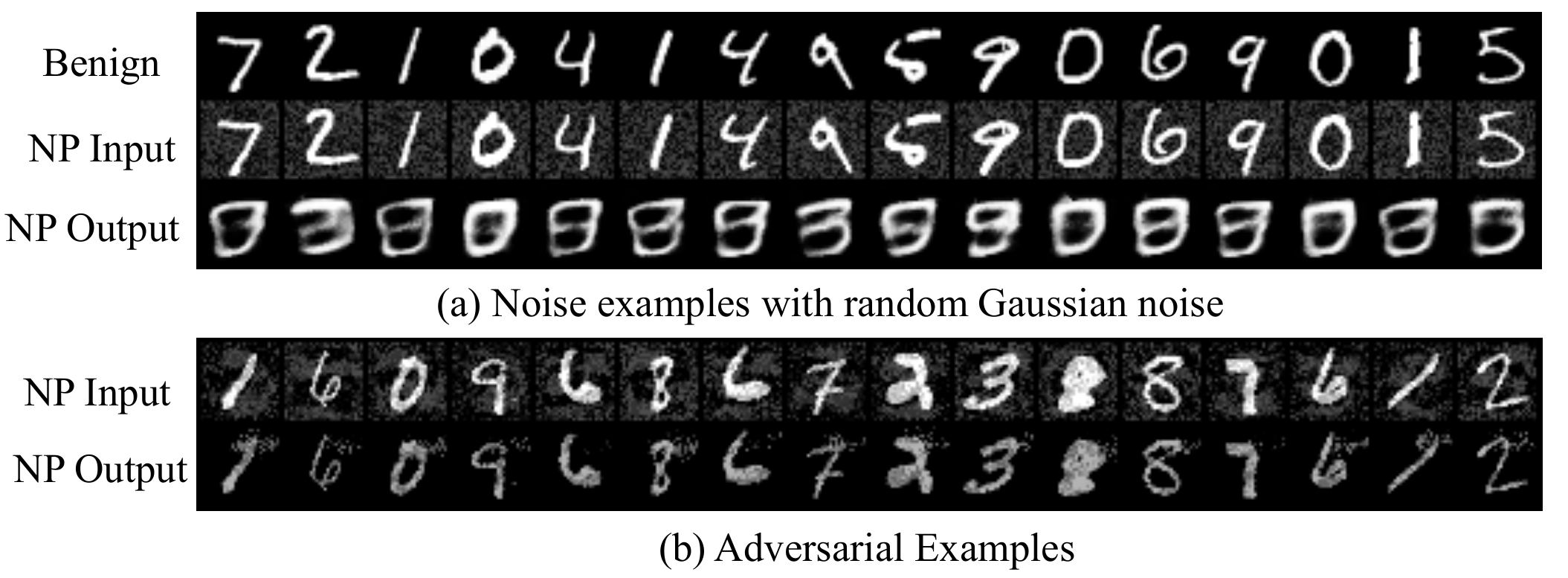}
    \caption{The reconstruction of pre-trained NP. Given noise examples with random Gaussian noises (a) and adversarial examples generated by PGD attack (b) under $L_\infty=0.2$ as the input of the benign pre-trained NP model, the adversarial examples could be reconstructed while noised examples could not.}
    \label{fig_motivation}
\end{figure}

\subsection{Discussion}
\label{sec:analyses}
As the adversarial examples in our NP-Attack is sampled and reconstructed through the decoder of the benign pre-trained NP, we study on the examples reconstructed by NP. Figure \ref{fig_motivation} shows that the adversarial examples and benign ones could both be reconstructed by benign pre-trained NP, sharing the same decoder, optimized with different latent variables. However, noised examples (added with slight Gaussian noises) could absolutely not be reconstructed by such NP. This phenomenon contributes to the query efficiency of NP-Attack compared to $\mathcal{N}$Attack, as the appliance of NP filters the noised examples away, the distributions modeled by NP are with higher percentages of adversarial examples, which keep consistency with the Figure \ref{fig:intro}.

Moreover, we discuss the difference between NP and other auto-encoder models, like VAE \cite{kingma2013auto}. Note that the variables in NP are different from VAE. In VAE, the latent variable $z$ is on the whole image set, which means a different sample would refer to the reconstruction of a different image. In NP, the distribution of $z$ is modeled independently on one image. So the optimization of $z$ in VAE to model adversarial distribution could lead to a collapse on distributions with high possibility. On the contrary, the Gaussian distribution of $z$ in NP could be optimized independently on different images towards adversarial distribution. Moreover, NP has another variable $r$, which is also modeled on one image and different from $z$, bringing more flexibility. 

\section{Experiment}
In this section, we evaluate our method on three benchmark image classification datasets MNIST \cite{L1998Gradient}, CIFAR10 \cite{cifar10} and ImageNet \cite{deng2009imagenet}. We compare our proposed NP-Attack with several score-based black-box attack techniques with regard to distortion ($\epsilon$), average query count (\textit{i.e.} the number of evaluations on black-box model) and attack success rate (ASR). We firstly provide a comprehensive understanding of NP-Attack where we test the performance of our NP-Attack under different experimental settings. After that, we conduct untargeted and targeted attacks on benchmark datasets MNIST, CIFAR10 and ImageNet to show the superiority of our method in reducing query counts.  

\textbf{Experimental Setups}
For MNIST, a MLP model is trained with three fully connected layers, achieving 98.50\% accuracy. For CIFAR10, the WideResNet \cite{zagoruyko2016wide} is adopted, achieving 94.10\% accuracy. For ImageNet, we use the pre-trained Inception-V3 model \cite{szegedy2016rethinking}, achieving 78\% accuracy. The baseline query based black-box techniques are ZOO \cite{chen2017zoo}, AutoZOOM \cite{tu2019autozoom}, QL \cite{ilyas2018black} and $\mathcal{N}$Attack \cite{li2019nattack}. The results of those compared methods are reproduced by the code released in the original paper with default settings. 

\textbf{NP Pre-training} For NP-Attack, NP models are pre-trained on the train set of MNIST, CIFAR10 and ImageNet respectively which takes around 5, 45 and 72 hours on a single NVIDIA GTX 1080 TI GPU. The dimensions of $r$ and $z$ are both 128. Regarding the computational overhead of the pre-training, we would like to point out that the pre-training of NP is over the whole dataset while the black-box attack is performed on a single image per time. The pre-trained model is used for free to perform black-box attack, when the pre-training is complete. Thus, the pre-training is a off-line operation, which is totally separated from the on-line black-box attack. The overhead of pre-training will not affect the black-box attack phase.

\subsection{Empirical Understanding of NP-Attack}
\label{sec:empirical understanding}
We evaluate the performance of NP-Attack under different experimental settings, including various sample sizes, maximum distortion, and optimization methods (optimizing $r$, $z$ or both). Experiments are conducted on MNIST with 200 correctly classified images selected from the test set. 

\textbf{Optimization Method} 
We first investigate the performance of our NP-Attack with different optimization methods, \textit{i.e.}, optimizing over the variable $r$, $z$ or both. The experimental results are summarized in Table \ref{table:optimization}. The $L_\infty$ distortion is set to 0.2 for MNIST and 0.05 for CIFAR10. Optimizing $r$ is more query efficient, for the reason that $r$ is modeled independent for different pixels in NP while $z$ is modeled as a same Gaussian distribution over all pixels in one image. NP-Attack-R and NP-Attack-Z show the upper and lower bound of our NP based attack. The adversarial examples generated by different optimization methods are shown in Appendix \ref{sec: ADV mnist}. 

\begin{table}[!htbp]
\begin{center}
\caption{Evaluation of NP-Attack under untargeted setting by optimizing the variables $r$, $z$ or both on 200 correctly classified images from MNIST and CIFAR-10.}
\label{table:optimization}
\begin{tabular}{l|ccc|ccc}
\hline\noalign{\smallskip}
\multirow{2}*{Attack Method} & \multicolumn{3}{c|}{MNIST} & \multicolumn{3}{c}{CIFAR10}\\
\cline{2-7}
 & ASR & $L_2$ Dist & Query Count & ASR & $L_2$ Dist & Query Count \\
\noalign{\smallskip}
\hline
\noalign{\smallskip}
NP-Attack-R & 100\% & 3.07 & \bf{1,190} & 100\% & 1.75 & \bf{96} \\
NP-Attack-Z & 100\% & 3.55 & 1,665 & 100\% & 1.68 & 150 \\
NP-Attack-RZ & 100\% & 3.65 & 1,460 & 100\% & 1.80 & 98 \\
\hline
\end{tabular}
\end{center}
\end{table}

\begin{figure}[!htbp]
    \centering
    \includegraphics[width=12cm]{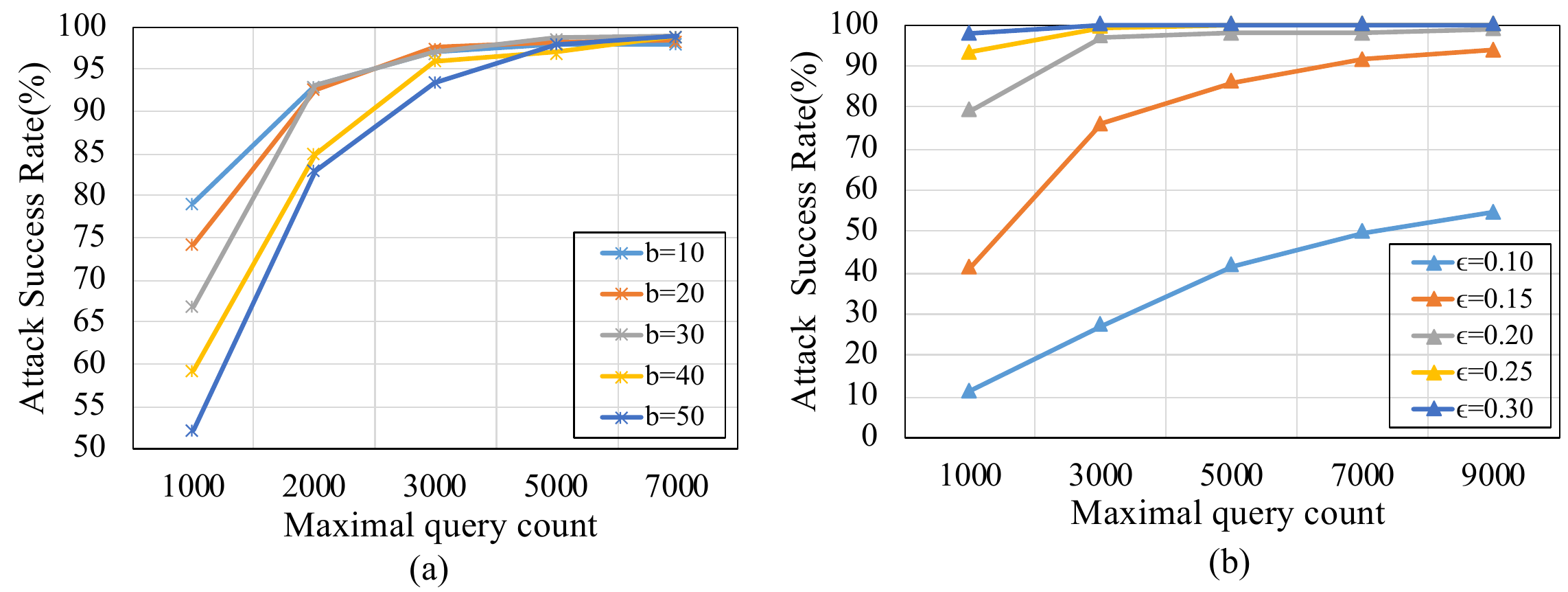}
    \caption{Change of ASRs when limiting the query counts on MNIST with NP-Attack-R. (a) Different curves represent performance under different sample sizes. (b) Different curves represent restricting different maximal distortion.}
    \label{fig: ablation study}
\end{figure}

\textbf{Sample Size}
We test the sample size $b \in \{10, 20, 30, 40, 50\}$ and set the $L_\infty$ distortion to 0.2. Noted that query count is linearly related to sample size, as $Q = T \times b$, where $T$ represents iteration count and $Q$ represents query count. The changes of ASR with the maximal query count are plotted in Figure \ref{fig: ablation study}(a). It shows that our NP-Attack could achieve over 90\% ASR with different sample sizes when the maximal query count is larger than 7000. The ASR, $L_2$ distortion and average query count when the maximal query count is 7000 are reported in Table \ref{table:sample size}. We can see that, with a larger sample size, the iteration $T$ is reduced along with larger average query counts. However, we can get higher ASRs and smaller $L_2$ distortion by increasing the sample size, illustrating a more accurate estimation of the perturbation direction.

\textbf{Maximal Distortion}
It is a trade-off between the distortion and query counts in black-box attack. We set the sample size $b=10$ and test the $L_\infty$ distortion $\epsilon \in \{0.10, 0.15, 0.20, 0.25, 0.30\}$. The change of ASRs with various maximal query counts is shown in Figure \ref{fig: ablation study}(b). 
When $\epsilon=0.3$, our NP-Attack achieves 98\% ASR with only 1000 maximal query counts. With the 5000 query counts, NP-Attack can achieve nearly 100\% ASR when $L_\infty$ is larger than 0.2. 

% \begin{table}[!t]
% \begin{center}
% \caption{Evaluation of NP-Attack-R under untargeted setting on MNIST with different sample sizes. The maximal query count is 7000 and $L_\infty$ distortion is restricted to 0.2.}
% \label{table:sample size}
% \begin{tabular}{p{2.5cm}p{3cm}p{1.5cm}p{1.5cm}p{2.5cm}}
% \hline\noalign{\smallskip}
% Sample Size (b) & Iteration Nums (T) & ASR & $L_2$ Dist & Avg Query Count \\
% \noalign{\smallskip}
% \hline
% \noalign{\smallskip}
% 10 & 700 & \bf{97.01\%} & 3.0146 & \bf{1,439} \\
% 20 & 350 & 94.03\% & 2.5684 & 1,487 \\
% 30 & 233 & 94.53\% & 2.4082 & 1,577 \\
% 40 & 175 & 92.54\% & 2.3279 & 1,657 \\
% 50 & 140 & 90.55\% & \bf{2.2694} & 1,869 \\
% \hline
% \end{tabular}
% \end{center}
% \end{table}

\begin{table}[!t]
\begin{center}
\caption{Evaluation of NP-Attack-R under untargeted setting on MNIST with different sample sizes. The maximal query count is 7000 and $L_\infty$ distortion is restricted to 0.2.}
\label{table:sample size}
\begin{tabular}{p{2.5cm}p{3cm}p{1.5cm}p{1.5cm}p{2.5cm}}
\hline\noalign{\smallskip}
Sample Size (b) & Iteration Nums (T) & ASR & $L_2$ Dist & Avg Query Count \\
\noalign{\smallskip}
\hline
\noalign{\smallskip}
10 & 700 & 98.01\% & 4.3707 & \bf{723} \\
20 & 350 & 98.51\% & 3.9478 & 786 \\
30 & 233 & \textbf{99.00\%} & 3.6675 & 940 \\
40 & 175 & \textbf{99.00\%} & 3.4675 & 1,074 \\
50 & 140 & \textbf{99.00\%} & \bf{3.3277} & 1,203 \\
\hline
\end{tabular}
\end{center}
\end{table}

\subsection{Evaluation on MNIST and CIFAR10}
\label{sec:evaluation on MNIST and CIFAR10}
For both MNIST and CIFAR10, we randomly select 1000 correctly classified images from the test set for untargeted attack. For targeted attack, 100 correctly classified images are selected from test set, for each image the target labels are set to the other 9 classes and a total 900 attacks are performed. In our experiments, the $L_\infty$ distortion is restricted to 0.2 for MNIST and 0.05 for CIFAR10 following the setting of \cite{ilyas2018prior}. For untargeted attack, the maximal iteration is $T=900$ and sample size is $b=30$ while for targeted attack $T=2000$, $b=50$. The learning rate is $\eta=0.01$ in default setting. A total query count is obtained by multiplying the number of iterations and the query count per iteration.

Note that the query count per iteration varies in different black-box attack techniques. ZOO uses the parallel coordinate-wise estimation with a batch of 128 pixels, resulting in 256 query counts per iteration. In AutoZOOM, the attack stages could be divided into initial attack success and post-success fine-tuning. In each iteration, the number of random vector is set to 1 at the first stage to find the initial attack success examples and then set to $q$ at the second stage to reduce the distortion at the same level with other techniques. Thus the query count for AutoZOOM is $q+1$ per iteration. For QL, $\mathcal{N}$Attack and our NP-Attack, query count in each iteration is the sample size $b$. For a fair comparison, we set the same sample size for these three NES algorithm based methods. Besides, both ZOO and AutoZOOM are optimization-based methods which quickly attack the model successfully and generate adversarial examples with quite large distortion and continuously perform post-success fine-tuning to reduce the distortion. We report the final fine-tuning query count and the distortion after fine-tuning for ZOO and AutoZOOM.

\begin{table}[tb]
\begin{center}
\caption{Adversarial evaluation of black-box attacks on MNIST.}
\label{table:untargeted attacks mnist}
\begin{tabular}{c|cccc|cccc}
\hline\noalign{\smallskip}
\multirow{2}*{Attack Method} & \multicolumn{4}{c|}{Untargeted Attack} & \multicolumn{4}{c}{Targeted Attack}\\
\cline{2-9}
 & ASR & $L_2$ & $L_\infty$ & Query Count & ASR & $L_2$ & $L_\infty$ & Query Count \\
\noalign{\smallskip}
\hline
ZOO & 100\% & 1.12 & 0.21 & 107,264 & 100\% & 1.64 & 0.29 & 128,768 \\
AutoZOOM-BiLIN & 100\% & 2.01 & 0.50 & 9,129 & 99.89\% & 2.78 & 0.62 & 9,401 \\
AutoZOOM-AE& 100\% & 2.62 & 0.67 & 10,202 & 99.89\% & 3.74 & 0.87 & 10,380  \\
QL & 96.62\% &3.38 & 0.20 & 2,549 & 99.67\% & 3.09 & 0.20 & 2,693 \\
$\mathcal{N}$Attack & 95.09\% & 2.14 & 0.20 & 4,357 & 98.22\% & 3.33 & 0.20 & 5,981 \\
\hline
NP-Attack-R(Ours) & 100\% & 3.09 & 0.20 & \bf{1,226} & 100\% & 3.72 & 0.20 & 2,693\\
NP-Attack-Z(Ours) & 99.90\% & 3.55 & 0.20 & 1,680 & 100\%  & 3.94 & 0.20 & \bf{2,605} \\
\noalign{\smallskip}
\hline
\end{tabular}
\end{center}
\end{table}

\begin{table}[t]
\begin{center}
\caption{Adversarial evaluation of black-box attacks on CIFAR10.}
\label{table:untargeted attacks cifar10}
\begin{tabular}{c|cccc|cccc}
\hline\noalign{\smallskip}
\multirow{2}*{Attack Method} & \multicolumn{4}{c|}{Untargeted Attack} & \multicolumn{4}{c}{Targeted Attack}\\
\cline{2-9}
 & ASR & $L_2$ & $L_\infty$ & Query Count & ASR & $L_2$ & $L_\infty$ & Query Count \\
\noalign{\smallskip}
\hline
ZOO & 100\% & 0.12 & 0.02 & 208,384 & 99.52\% & 0.19  & 0.02 & 230,912 \\
AutoZOOM-BiLIN & 100\% & 1.56 & 0.15 & 8,113 & 100\% & 2.13 & 0.21 & 8,266 \\
AutoZOOM-AE& 100\% & 1.88 & 0.16 & 7,113 & 100\% & 2.78 & 0.24 & 8,217\\
QL & 98.40\% & 1.91 & 0.05 & 857 & 99.55\% & 2.11 & 0.05 & 616  \\
$\mathcal{N}$Attack & 99.89\% & 2.61 & 0.05 & 183 & 100\% & 2.61 & 0.05 & 1,151 \\
\hline
NP-Attack-R(Ours)& 100\% & 1.74 & 0.05 & \bf{94} & 100\% & 1.85 & 0.05 & \bf{589}\\
NP-Attack-Z(Ours) & 100\% & 1.67 & 0.05 & 144 & 100\% & 1.78 & 0.05 & 936 \\
\noalign{\smallskip}
\hline
\end{tabular}
\end{center}
\end{table}

The experimental results are shown in Tables \ref{table:untargeted attacks mnist} and \ref{table:untargeted attacks cifar10}. We report the ASRs, $L_2$ distortion, $L_\infty$ distortion and average query counts. As mentioned in Section \ref{sec:empirical understanding}, performance of NP-Attack-RZ is between NP-Attack-R and NP-Attack-Z. Thus, in the following experiments, we only show the results of NP-Attack-R and NP-Attack-Z to benchmark the upper and lower bound of our NP based attack.  Compared to ZOO, our NP-Attack-R can reduce the query count by 98.86\% and 99.95\% on MNIST and CIFAR10 respectively under untargeted attack setting, while for targeted attack, NP-Attack-R reduces the query count by 96.91\% and 99.74\% on MNIST and CIFAR10. The comparison shows that though ZOO achieves 100\% ASR, it is far from query efficient, requiring over 100,000 queries to generate considerable adversarial examples, implying high costs in computation and time. When it comes to AutoZOOM, it could significantly reduce the query count by proposing an adaptive random gradient estimation strategy. AutoZOOM-BiLIN and AutoZOOM-AE leverage a simple bilinear resizer or auto-encoder as the decoder respectively to reduce the dimension of adversarial perturbations. AutoZOOM based method can attack the model successfully with quite fewer initial success query counts, but the distortion is unacceptable at initial success. It still requires much more queries to fine-tuning. Compared to AutoZOOM, our NP-Attack methods outperform not only in query counts but also in the $L_\infty$ distortion. QL utilizes NES algorithm to estimate the gradient, thus the performance of such technique hinges on the quality of the estimated gradient. QL shows its superiority in targeted attack for the reason that gradient-based methods are easier to find the targeted direction. In contrast, $\mathcal{N}$Attack and our NP-Attack both utilize NES to estimate the adversarial distribution. Compared to QL, our NP-Attack-R could achieve much better results in untargeted attack and comparable results in targeted attack. Compared to $\mathcal{N}$Attack, our NP-Attack could achieve higher ASRs and fewer query counts under both targeted or untargeted settings, due to the outstanding distribution modeling capacity of the NP model. In general, NP-Attack obtains the best trade-off between query counts and distortion. More experimental results on various architectures are reported in Appendix \ref{sec: additional architectures}.

\subsection{Evaluation on ImageNet}
\label{sec:evaluation on ImageNet}

The dimension of the images in ImageNet is relatively larger, which requires more queries to generate adversarial examples. We randomly select 100 correctly classified images from the test set to perform untargeted and targeted black-box attacks on ImageNet. For each image in targeted attack, a random label except the true one out of 1000 classes is selected to serve as the target.
The $L_\infty$ distortion is restricted to 0.05 following the setting of \cite{ilyas2018black,ilyas2018prior}, $T = 600$, $b = 100$ and the learning rate $\eta = 0.005$. 
To improve the query efficiency on images with large sizes, ZOO and AutoZOOM utilize the techniques such as hierarchical attack and compressing dimension of attack space. For our NP-Attack, we also perform such compression. To be specific, we resize the images to 32$\times$32$\times$3 to train the NP model, then we add perturbations on each 32$\times$32$\times$3 patch, which is cut independently from the original images.  Although using the attack dimension reduction, our NP-Attack method still outperforms.

The experimental results are summarized in Table \ref{table:attacks on ImageNet}. Due to the large image sizes, ZOO suffers from low ASRs and tremendous model evaluations especially for targeted attack. For other three compared attacks, they achieve 100\% ASRs at the cost of over 10,000 queries in targeted attack and over 1,000 queries in untargeted attack. While for our NP-Attack, NP-Attack-R can achieve 100\% ASR with only 867 queries, yielding a 94.45\% query reduction ratio compared to ZOO in untargeted attack. In targeted attack, NP-Attack-R and NP-Attack-Z get over 98\% ASRs and reduce the query counts to 8,001 and 11,383 respectively, which exceed all the compared methods. This demonstrates that our NP-Attack can scale to ImageNet set. The generated adversarial examples by NP-Attack-R are shown in Appendix \ref{sec: ADV Imagenet}.

\begin{table}[ht]
\begin{center}
\caption{Adversarial evaluation of black-box attacks on ImageNet.}
\label{table:attacks on ImageNet}
\begin{tabular}{c|ccc|ccc}
\hline\noalign{\smallskip}
\multirow{2}*{Attack Method} & \multicolumn{3}{c|}{Untargeted Attack} & \multicolumn{3}{c}{Targeted Attack}\\
\cline{2-7}
 & ASR & $L_2$ Dist & Query Count & ASR & $L_2$ Dist & Query Count \\
\noalign{\smallskip}
\hline
\noalign{\smallskip}
ZOO & 90\% & 1.20 &  15,631 & 78\% & 3.43 & 2.11x10$^6$ \\
AutoZOOM-BiLIN & 100\% & 9.34 & 3,024 & 100\% & 11.26 & 14,228 \\
QL & 100\% & 17.72 & 3,985 &  100\% & 17.39 & 33,360\\
$\mathcal{N}$Attack & 100\% & 24.01 & 2,075 & 100\%  & 24.14 & 14,229\\
Bandits & 100\% & -- & 1,165 & 100\% & -- & 25,341 \\
\hline
NP-Attack-R(Ours) & 100\% & 10.96 & \bf{867} & 98.02\% & 14.38 & \bf{8,001} \\
NP-Attack-Z(Ours) & 96.04\% & 12.37 & 1,236 & 98.02\% & 14.60 & 11,383  \\
\hline
\end{tabular}
\end{center}
\end{table}

\section{Conclusions}
In this paper, we focus on improving the query efficiency in black-box attack by modeling the high-level distribution of adversarial examples. By considering the structure information of pixels in one image rather than individual pixels, which is realized by a decoder of benign pre-trained Neural Process model, we propose the Neural Process-based black-box attack (NP-Attack) to greatly reduce the required query complexity. 
Evaluated on MNIST, CIFAR10 and ImageNet with regard to query count and attack success rate, our proposed NP-Attack achieves the state-of-the-art results, showing its efficiency and superiority under the black-box adversarial attack setting. Moreover, when pre-training the NP model on adversarial image examples instead of benign examples, we believe the query efficiency could be further improved, which is left for our future work.
 
% ---i- Bibliography ----

\section{Acknowledgement}
This work is supported in part by the National Key Research and Development Program of China under Grant 2018YFB1800204, the National Natural Science Foundation of China under Grant 61771273, the R\&D Program of Shenzhen under Grant JCYJ20180508152204044, and the project 'PCL Future Greater-Bay Area Network Facilities for Large-scale Experiments and Applications (LZC0019)'. We also thanks for the GPUs supported by vivo and Rejoice Sport Tech. co., LTD.

\appendix

\section{The Structure of ANP in NP-Attack}\label{appendix:structure}
Here, we show the detailed structure of the pre-trained ANP model on CIFAR10.
\begin{table}[!htbp]
    \centering
    \caption{Structure of the deterministic part of encoder $h$. (The input of Linear1 includes the pixel position: 3072 (32 $\times$ 32 $\times$ 3) $\times$ 3, the pixel value in RGB respectively: 3072 $\times$ 1, thus with a size of 3072 $\times$ 4 in total.)}
    % \vspace{2mm}
    \begin{tabular}{|c c c c|}
    \hline
    Layer & Input & Output & Activation function\\
    \hline\hline
    Linear1 & 3072$\times$4 & 3072$\times$128 & ReLU \\
    Linear2 & 3072$\times$128 & 3072$\times$128 & ReLU \\
    Linear3 & 3072$\times$128 & 3072$\times$128 & ReLU \\
    SelfAtt & 3072$\times$128 & 3072$\times$128 & -\\
    CrossAtt & 3072$\times$128 & 3072$\times$128 & - \\
    Linear4 & 3072$\times$128 & 3072$\times$128 & - \\  
    \hline
    \end{tabular}
    %  \vspace{3mm}
    \label{tab:anp_ed}
    % \vspace{2mm}   
\\~\\~

    \caption{Structure of the sampled latent part of encoder $h$.}
    % \vspace{2mm}
    \begin{tabular}{|c c c c|}
    \hline
    Layer & Input & Output & Activation function\\
    \hline\hline
    Linear1 & 3072$\times$4 & 3072$\times$128 & ReLU \\
    Linear2 & 3072$\times$128 & 3072$\times$128 & ReLU \\
    Linear3 & 3072$\times$128 & 3072$\times$128 & ReLU \\
    SelfAtt & 3072$\times$128 & 3072$\times$128 & -\\
    Mean & 3072$\times$128 & 3072$\times$128 & - \\
    Linear4 & 3072$\times$128 & 3072$\times$128 & - \\  
    \hline
    \end{tabular}
    % \vspace{3mm}
    \label{tab:anp_ed}
    % \vspace{2mm}  
\\~\\~

    \caption{Structure of the latent part of decoder $g$. (The input of Linear1 includes the deterministic path: 3072 $\times$ 128, the sampled latent path: 3072 $\times$ 128 and the target pixel position: 3072 $\times$ 3, thus with a size of 3072 $\times$ 259 in total.)}
    % \vspace{2mm}
    \begin{tabular}{|c c c c|}
    \hline
    Layer & Input & Output & Activation function\\
    \hline\hline
    Linear1 & 3072$\times$259 & 3072$\times$128 & ReLU \\
    Linear2 & 3072$\times$128 & 3072$\times$128 & ReLU \\
    Linear3 & 3072$\times$128 & 3072$\times$128 & ReLU \\
    Linear4 & 3072$\times$128 & 3072$\times$1 & ReLU \\
    \hline
    \end{tabular}
    % \vspace{-3mm}
    \label{tab:anp_de}
\end{table}

\section{More Optimization Options of NP-Attack}
\label{NPattack_branches}
Here we show the details of the other two optimization options, NP-Attack-Z and NP-Attack-RZ. 
\subsubsection{NP-Attack-Z} 
In this case, we fix $r$ and optimize on the distribution of $z$, in order to sample $z$ with adversary. Under the guidance of score outputs in each query, to be simplified, we just optimize the predictive mean $\mu$, keeping $\sigma$ fixed. The optimization of our NP-Attack is inherited from NES, which could enhance query efficiency compared with those vector-wise gradient estimation strategies.
Omitting the fixed $r$ while inheriting \textit{Decoder $g$} from the pre-trained NP, the estimated loss function in our NP-Attack could be considered as $L = \mathbb{E}_{\mathcal{N}(z \mid \mu, \sigma^2)}l(g(z))$.
The optimization of $\mu$ is implemented by adding some random Gaussian noises and further using NES to estimate the gradient. Those added Gaussian noises are of the same variance with $z$ in pre-trained NP. The reason comes as that a same variance could keep $z$ of a simpler representation for variance after adjustment in each iteration, which benefits the optimization later. The optimization function could be computed as: $\mu_{t+1} \leftarrow \mu_t - \eta \nabla_{\mu}L\mid_{\mu_t},$ where $\eta$ denotes a learning rate. To guarantee such optimization being more accurate, in practice the optimization is over a mini-batch of sample size $b$, which means we add $b$ random perturbations independently during each iteration and operate on those outputs. So the updating operation is on the average values instead:
\begin{equation}
    \mu_{t+1} \leftarrow \mu_t - \frac{\eta}{b} \sum_{i=1}^{b} l(g(z_{i}))\nabla_{\mu_t}\log \mathcal{N}(z_{i} \mid \mu_i, \sigma^2).
\end{equation}
Given that $\mu_i = \mu_t + p_i \sigma, z_i = \mu_i + p_i \sigma = \mu_t + 2p_i \sigma$, where $p_i$ is sampled from the standard Gaussian distribution $\mathcal{N}(0,I)$, and $\sigma$ is multiplied to keep a simple representation for variance, 
$\nabla_{\mu_t} \log \mathcal{N}(z_{i} \mid \mu_i, \sigma^2) \varpropto \sigma^{-1} p_i.$

\subsubsection{NP-Attack-RZ}
Similarly, we propose NP-Attack-RZ as a third choice by optimizing both $r$ and $z$ simultaneously. The operation is the same with NP-Attack-R/Z as shown in Algorithm \ref{alg:NP-Attack}.

\section{The Visual Comparison of Adversarial Examples by Different NP-Attack Versions.}\label{sec: ADV mnist}
\begin{figure}[!htbp]
    \centering
    \includegraphics[width=0.5\textwidth]{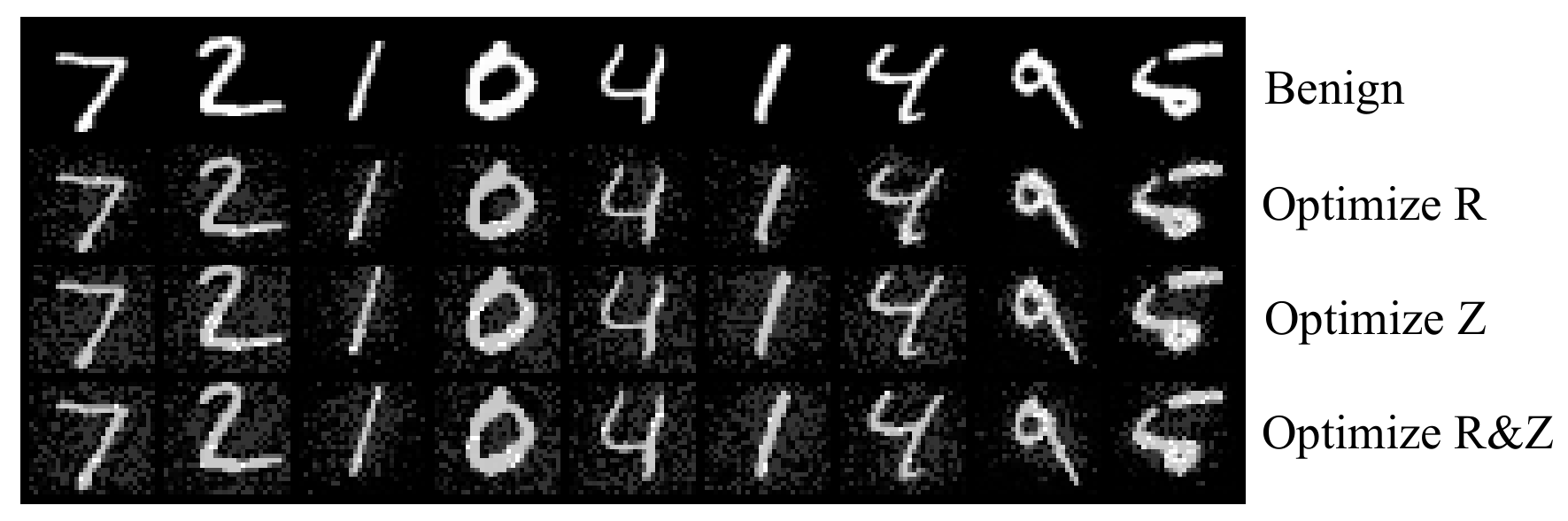}
    \caption{Adversarial examples of different optimization methods (NP-Attack-R/Z/RZ). }
    \label{fig:adv of diffent optimization}
\end{figure}

We show the adversarial examples generated by NP-Attack-R/Z/RZ in Figure \ref{fig:adv of diffent optimization}. We observe that adversarial perturbations generated by NP-Attack-R are centered around the main digits while perturbations generated by NP-Attack-Z are scattered on the background. Furthermore, the location of adversarial perturbations in NP-Attack-RZ is somehow in a moderate degree between NP-Attack-R and NP-Attack-Z. 

\section{Evaluation on Additional Architectures}
\label{sec: additional architectures}
We additionally do experiments under $\epsilon=0.031$ on CIFAR-10 on various architectures including ResNet18 \cite{he2016deep}, VGG16 \cite{simonyan2014very} and WideResNet \cite{zagoruyko2016wide}. We compare NP-Attack-R with the state-of-the-art $\mathcal{N}$Attack here and report the attack success rate (ASR) and average queries in Table \ref{table:attacks on other architectures}. We conduct untargeted attacks with the same experimental setting of Section \ref{sec:evaluation on MNIST and CIFAR10}. The experimental results demonstrated that our method still outperforms towards various architectures.

\begin{table}[ht]
\begin{center}
\caption{Adversarial evaluation of untargeted black-box attacks on CIFAR-10 on various architectures.}
\label{table:attacks on other architectures}
\begin{tabular}{c|ccc|ccc}
\hline\noalign{\smallskip}
\multirow{2}*{Attack Method} & \multicolumn{3}{c|}{ASR} & \multicolumn{3}{c}{Avg Queries} \\
 & ResNet18 & VGG16 & WRN  & ResNet18 & VGG16 & WRN \\
\noalign{\smallskip}
\hline
\noalign{\smallskip}
$\mathcal{N}$Attack &  96.79\% & \textbf{100\%} & 99.89\% & 311 & 430 & 335 \\
NP-Attack-R(Ours) & \textbf{100\%} & \textbf{100\%} & \textbf{100\%} & \textbf{185} &  \textbf{294} & \textbf{196}\\
\hline
\end{tabular}
\end{center}
\end{table}

\section{The Generated Adversarial Examples on ImageNet by NP-Attack}\label{sec: ADV Imagenet}
We show the adversarial examples generated by our NP-Attack-R here. It is demonstrated that though we perturb each 32$\times$32$\times$3 patch in the original images, the perturbation generated by our method is still imperceptible.

\begin{figure}[!htbp]
    \centering
    \includegraphics[width=9cm]{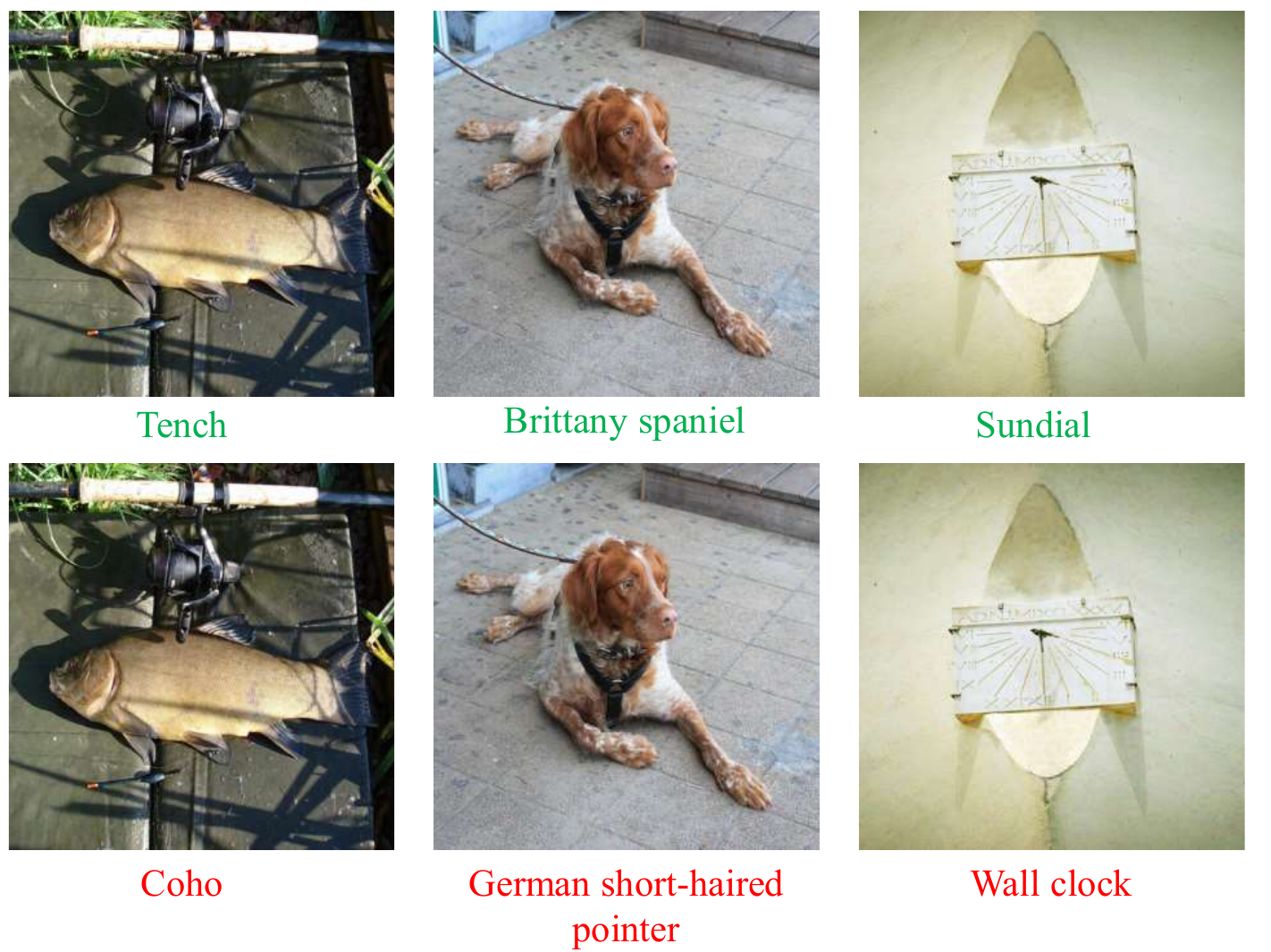}
    \caption{Adversarial example of ImageNet generated by NP-Attack-R on untargeted attack bounded by $L_\infty$=0.05. Top row shows the original images and the true labels while the bottom row are the adversarial examples with predicted label. }
    \label{fig: ADV Imagenet}
\end{figure}

\bibliographystyle{splncs04}
\bibliography{egbib}

\end{document}